\documentclass[11pt,a4paper]{article}
\usepackage{microtype}
\usepackage[hyperref]{emnlp-ijcnlp-2019}
\usepackage{times}
\usepackage{latexsym}
\usepackage[normalem]{ulem}
\usepackage{graphicx}

\usepackage{url}

\aclfinalcopy 

\title{On the Importance of Delexicalization for Fact Verification}

\author{Sandeep Suntwal\thanks{~~Equal Contribution}, Mithun Paul\footnotemark[1], Rebecca Sharp, Mihai Surdeanu\\
  University of Arizona, Tucson, Arizona, USA\\
  {\tt \{sandeepsuntwal,mithunpaul,bsharp,msurdeanu\}@email.arizona.edu}
}  

\date{}

\begin{document}
\maketitle
\begin{abstract}

While neural networks produce state-of-the-art performance in many NLP tasks, they generally learn from lexical information, which may transfer poorly between domains. Here, we investigate the importance that a model assigns to various aspects of  data while learning and making predictions, 
    specifically, in a recognizing textual entailment (RTE) task. By inspecting the attention weights assigned by the model, we confirm that most of the weights are assigned to noun phrases. To mitigate this dependence on lexicalized information, we experiment with two strategies of masking.  
First, we replace named entities with their corresponding semantic tags along with a unique identifier 
to indicate lexical overlap between claim and evidence.  Second, we similarly replace other word classes in the sentence (nouns, verbs, adjectives, and adverbs)  with their super sense tags \citep{ciaramita2003supersense}.
Our results show that, while performance on the in-domain dataset remains on par with that of the model trained on fully lexicalized data, it improves considerably when tested out of domain. 
For example, the performance of a state-of-the-art RTE model trained on the masked Fake News Challenge \citep{pomerleau2017fake} data and evaluated on Fact Extraction and Verification  \cite{thorne2018fever} data improved by over 10\% in accuracy score compared to the fully lexicalized model.
\end{abstract}

\section{Introduction}

Neural networks (NNs)  play a key role in most modern natural language processing (NLP) systems, obtaining state-of-the-art (SOA) performance~\citep{devlin2018bert, sun2018improving,bohnet2018morphosyntactic} in many complex tasks, e.g., recognizing textual entailment~\citep{kim2018semantic}, fake news detection~\citep{baird2017talos} and fact verification~\citep{nie2018combining}.

However, these models depend heavily on lexical information that may transfer poorly between different domains. For example, in early experiments in the fact verification space, we observed that out of all the 34 statements containing the phrase ``American Author,'', 31 (91\%) belonged to one class label. Such information could be meaningful say, in the literature domain, but transfers poorly to other domains such as science or entertainment. 

In this work we aim to: (a) understand and estimate the importance that a neural network assigns to various aspects of the data while learning and making predictions, and (b) learn how to control for unnecessary lexicalization. Here we focus on the recognizing textual entailment (RTE) task \citep{dagan2013recognizing}, and its application to fact verification \citep{thorne2018fever, pomerleau2017fake}.

RTE is the task of determining if one piece of text can be plausibly inferred from another. In the Fact Extraction and Verification (FEVER) shared task \cite{thorne2018fever}, the RTE module was used determine if a given set of evidence sentences, when compared with the claim provided, can be classified as \textit{supports, refutes}, or \textit{not enough information}.
In this context, the contributions of our work are:

{\flushleft {\bf (1)}} To verify that models trained on lexicalized data transfer poorly, we implement a domain transfer experiment where a state-of-the-art RTE model ~\cite{parikh2016decomposable}  is trained on the FEVER data, and tested on the Fake News Challenge (FNC) \citep{pomerleau2017fake} 
dataset, and vice versa. As expected, even though this method achieves high accuracy when evaluated in the same domain, the performance in the target domain is poor, marginally above chance.

 {\flushleft {\bf (2)}} We perform an error analysis and examine the attention weights that the model  assigns when making incorrect predictions.
 
 With this analysis, we are able to confirm that most of the weight is assigned to POS tags of nouns or elements of noun phrases, which confirms our observation that these models anchor themselves on lexical information that is more likely to be domain dependent. 
 
{\flushleft {\bf (3)}} To mitigate this dependence on lexicalized information, we experiment with several strategies for delexicalization, i.e., where lexical tokens are replaced (or masked) with indicators of their class. While the technique of delexicalization/masking has been used before \citep[e.g.,]{zeman2008cross}, here we expand it by incorporating semantic information.

In particular, we first replace named entities with their corresponding semantic tags from Stanford's CoreNLP \citep{manning2014stanford}. 
To keep track of which entities are referenced between claim and evidence sentences, we extend these tags with unique identifiers. 
Second, we similarly replace other word classes in the sentence (common nouns, verbs, adjectives, and adverbs)  with their super sense tags \citep{ciaramita2003supersense}.

{\flushleft {\bf (4)}}  The evaluation of the proposed masking strategy on the two fact verification datasets indicates that,
while the in-domain performance remains on par with that of the model trained on the original, lexicalized data, it improves considerably when tested in the out-of-domain dataset. 
For example, the performance of a state-of-the-art RTE model trained on the masked FNC data and evaluated on FEVER data improved by over 10\% in accuracy score compared to the fully lexicalized model. Similarly, the model trained on the masked FEVER data and tested on FNC outperforms the lexicalized model by 4.7\% FNC score.
Thus our experiments demonstrate that our masking strategy is successful in mitigating the dependency on domain-specific lexical information.

All the software for our proposed approach is open-source and publicly available on GitHub at: \url{https://github.com/clulab/releases/tree/master/emnlp2019-masking}.

\section{Setup}

\subsection{RTE Method}
For all of our experiments we use the Decomposable Attention (DA) model \cite{parikh2016decomposable}, which consistently achieves near state-of-the-art performance on RTE tasks\footnote{This experiment was initially setup during the 2018 FEVER shared task which had provided Decomposable Attention as a strong baseline. At the completion of the shared task, most of the winning teams used another model \cite{chen2016enhanced}. However, the aim of this paper is not to outperform the state of the art performance, but rather to generalize to new domains without retraining.}. In particular, we use the AllenNLP\footnote{\url{https://github.com/allenai/allennlp}}
implementation of DA, which was provided by the FEVER task organizers.  

\begin{table}
    \centering
    \footnotesize
    \begin{tabular}{ccccc}
        Dataset & Support & Refute & Discuss & Unrelated \\
        \hline
        FNC    & 5,581 & 1,537  & 13,373 & 54,894 \\
        FEVER  &  86,701 & 36,441  & \multicolumn{2}{c}{42,305 (\textit{NEI})} \\  
    \end{tabular}
    \caption{Label distribution for the FEVER and FNC datasets.  We consider the \textit{agree} and \textit{disagree} FNC labels as equivalent to the \textit{support} and \textit{refute} labels in FEVER. The FNC \textit{not enough info (NEI)} label is listed below the more fine-grained \textit{discuss} and \textit{unrelated} FEVER labels.  }
    \label{tab:data}
\end{table}

\subsection{Datasets}
We use two distinct fact verification datasets for our experiments (see Table~\ref{tab:data} for a summary):

{\flushleft{\bf Fake News Challenge (FNC):}} 
The FNC dataset~\citep{pomerleau2017fake} contains four classes (\textit{agree}, \textit{disagree}, \textit{discuss}, and \textit{unrelated}) and has publicly available training (49,972 data points) and test partitions (25,413 data points). Each data point consists of a claim and a set of evidence sentences.  We split the training dataset into two partitions, with 40,904 records for training and 9,068 records for development.

{\flushleft{\bf Fact Extraction and Verification (FEVER):}} The FEVER \cite{thorne2018fever} dataset consists of 145,449 training data points, each of which has a claim and a set of evidences retrieved from Wikipedia using a baseline information retrieval (IR) module.
The FEVER claim-evidence pairs were assigned labels from three classes: \textit{supports}, \textit{refutes}, and \textit{not enough info (NEI)}.  

Even though the partition of the FEVER dataset that was used in the final shared task competition was released publicly, the gold test labels were not. Hence we used the development portion (19,998 data points) instead as our test partition. We created our own development partition by randomly dividing the training partition into 80\% for training (119,197 data points) and 20\% (26,252 data points) for development.  The evidence for data points that had the gold label of \textit{not enough info} can be retrieved (using a task-provided IR component) either by finding the nearest neighbor to the claim or randomly \cite{thorne2018fever}.

\subsection{Cross-domain Labels}
\label{sec:crossdomain}

In order to evaluate models in a cross-domain setting, we modified the label space of the source domain to match that of the target domain. This  also allows us to evaluate using the official scoring measures of the target domain.

In particular, when training on FEVER and testing on FNC, the data points in FEVER that belong to the class \textit{supports} were relabeled as \textit{agree}, and those in \textit{refutes} as \textit{disagree}. The data points belonging to the third class \textit{NEI} were divided into \textit{discuss} and \textit{unrelated} as follows.
In the FEVER dataset, two separate methods were used to retrieve the evidence sentences for the class \textit{NEI}: 1) sampling a sentence from the nearest page to the claim as evidence using their document retrieval component, and 2) sampling a sentence from Wikipedia uniformly at random. The evidence sentences retrieved using the nearest page technique were assigned the label \textit{discuss} (since it was more likely to be topically relevant to the claim), and the rest were assigned the label \textit{unrelated}. Also the distribution of labels was made similar to that of FNC. In particular, 16.86\% of the total 35,639 data points that originally belonged to the class NEI in FEVER were thus assigned the label \textit{discuss} and the rest (83.14\%) were assigned the label \textit{unrelated}. For the experiments in the opposite direction, i.e., when training on FNC and testing on FEVER, we collapsed the \textit{unrelated} and \textit{discuss} classes into a single class, \textit{not enough info}.

To investigate DA's reliance on lexical information, we first examine its word-level attention weights (Section \ref{attention_analysis}). 
Informed by these results, we propose several methods to mask the lexicalized data (Section \ref{masking_techniques}), and evaluate them in in-domain and  out-of-domain settings (Section \ref{sec:results}).

\section{Error Analysis of Cross-Domain Attention} \label{attention_analysis}

\begin{figure}
 \includegraphics[width=0.95\linewidth]{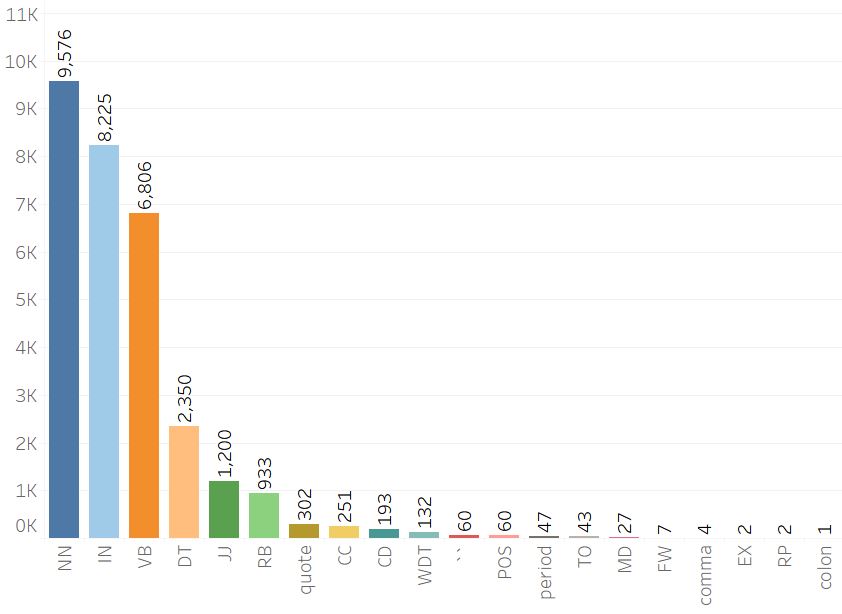}
    \vspace{-3mm}
    \caption{ Distribution of POS tags that were assigned the highest attention weights by DA for incorrectly classified cross-domain examples.}
  \label{fig:attention}
\vspace{-6mm}
\end{figure}

We use the word-level attention weights~\cite{bahdanau2014neural} learned by DA to perform  an error analysis in the cross-domain evaluation. 
In particular, we first trained two equivalent DA models, one on FEVER and the other on FNC. 
Next, we used both models to predict instances in FEVER development. For each data instance, we calculated the cumulative attention weights assigned by each of the two models to all evidence words. 
For each of the data instances that were {\em incorrectly} classified by the model trained out of domain (i.e., on FNC) and were {\em correctly} classified by the in-domain model, 
we  selected the words that were present in the set of top three words with the highest attention score according to the out-of-domain model, but were {\em not} present in the equivalent set produced by the in-domain model. Such words indicate potential overfitting of the out-of-domain model. 
Figure \ref{fig:attention} shows the distribution of part-of-speech (POS) tags for these words. The figure indicates that, for incorrectly classified examples in cross-domain, the higher attention weights were assigned to nouns.
Importantly, 43.10\% of these noun phrases in FEVER are named entities.

\section{Masking Techniques}\label{masking_techniques}

To mitigate the potential domain dependence introduced by these large attention weights, we propose several semantic masking techniques, which we compare with a deletion baseline.  Examples of each form of masking are shown in Table~\ref{masking_examples}.

\begin{table*}[t]
\begin{center}
\begin{tabular}{p{20mm}|p{55mm}|p{70mm}}

\textbf{Config.} & \textbf{Claim}& \textbf{Evidence} \\ \hline
Lexicalized & {With Singapore Airlines, the Airbus A380 entered commercial service.} & {The A380 made its first flight on 27 April 2005 and entered commercial service on 25 October 2007 with Singapore Airlines.}\\
\hline 
NE Deletion & {With  , the  entered commercial service.} & {The A380 made its  flight on  and entered commercial service on  with.}\\
\hline 
Basic NER  & {With \texttt{organization}, the \texttt{miscellaneous} entered commercial service.} & {The A380 made its \texttt{ordinal} flight on \texttt{date} and entered commercial service on \texttt{date} with \texttt{organization}.}\\
\hline 
OA-NER  & {With \texttt{organization-c1}, the \texttt{misc-c1} entered commercial service.} & {The A380 made its \texttt{ordinal-e1} flight on \texttt{ date-e1} and entered commercial service on \texttt{date-e2} with \texttt{organization-c1}.}\\
\hline 

\mbox{OA-NER+SS} Tags & {With \texttt{organization-c1}, the \texttt{artifact-c1} \texttt{motion-c1} commercial \texttt{act-c1} .} & {The A380 \texttt{stative} its \texttt{ordinal-e1 cognition-e1} on \texttt{date-e1 date-e2 date-e3} and \texttt{motion-c1} commercial \texttt{act-c1 on date-e4 date-e5 date-e6} with \texttt{organization-c1}.  
}\\

\end{tabular}
\end{center}

    \caption{ Example illustrating our various masking techniques, compared to the original fully lexicalized data. Note that the masking tags were generated with real-world (imperfect) tools. For example, ``Airbus A380" in the claim was correctly classified as \texttt{miscellaneous} by the NER tool, while ``A380" in the evidence was not, thus preventing us from taking advantage of the overlap. }
    \label{masking_examples}
\end{table*}

{\flushleft{\textbf{Named Entity (NE) Deletion Baseline:} }}
Lexical items which are tagged as named or numeric entities (NE) by CoreNLP's named entity recognizer (NER)~\citep{manning2014stanford} are deleted.  

{\flushleft{\textbf{Basic NER:}}}  Token sequences which are labeled as NEs are replaced by the corresponding label, e.g., \texttt{location, person}.

{\flushleft{\textbf{Overlap Aware NER (OA-NER)}: }} This technique additionally captures the \textit{lexical overlap} between the claim and evidence sentences with entity ids.  
That is, the first instance of a given entity in the claim is tagged with \texttt{c1}, where the \texttt{c} denotes the fact that it was found in the claim sentence (e.g., \texttt{person-c1}). Wherever this {\em same} entity is found later, in claim or in evidence, it is replaced with this unique tag. If an entity is found only in evidence, then it is denoted by an \texttt{e} tag. (e.g., \texttt{location-e3} would be the third location found only in the evidence).

For example, in the claim-evidence pair shown in Table~\ref{masking_examples}, when the named entity \textit{Singapore Airlines} appears in the claim it is replaced with \texttt{organization-c1}, since it is the first \texttt{organization} to appear in claim. 
The same id is used wherever the same entity is seen again, e.g., in the evidence sentence. However, the date \textit{27 April 2005} occurs only in the evidence, and hence it is replaced with \texttt{date-e1}.
Importantly, we create pseudo-pretrained embeddings for these new OA-NER-based tokens by adding a small amount of random Gaussian noise (mean 0 and variance of 0.1) to pre-trained embeddings~\citep{pennington2014glove} of the root word corresponding to the category (e.g., \textit{person}). Thus the embeddings of all the sub-tags, while being unique, are close to that of the root word.
{\flushleft{\textbf{OA-NER + Super Sense (SS) Tags}}}:
Super-sense tagging is a sequence modeling approach that annotates phrases with coarse WordNet senses~\citep{ciaramita2003supersense,miller1990introduction}. In this masking method, we not only replace named entities with their OA-NER tags, but also replace other lexical items with their corresponding super sense tags, if found. As with the OA-NER approach, the lexical overlap is also explicitly marked for all these tags with unique ids (see Table~\ref{masking_examples}).

\begin{table*}[ht]
\begin{center}
\begin{tabular}{p{22mm}|p{9mm}p{9mm}p{9mm}p{9mm}}
 & \multicolumn{4}{c}{Configuration} \\
 \hline
Train Domain & {FNC}& {FEVER}  & {FEVER} & {{FNC}} \\ 
Eval Domain & {FNC}& {{FNC}}  & {FEVER} & {{FEVER}} \\ \hline
Masking & & & & \\
\hline
Lexicalized &68.99\%& {48.86\%} &83.43\%& {41.16\%} \\
Deletion  &66.45\%& 40.23\% &75.34\%& 33.33\% \\
Basic NER &69.40\%& 46.27\% &76.23\%& 35.72\%\\
\textbf{OA-NER} &65.85\%& \textbf{53.59}\% &{82.31\%}& {46.47\%}\\
\textbf{OA-NER+SS} & 45.51\%& 46.71\% &75.26\%& {\bf 51.77\%}\\
\end{tabular}
\end{center}
    \caption{\label{crossdomain} Various masking techniques and their performance accuracies, both in-domain and out-of-domain.} \label{tab:results}

\end{table*}

\section{Results} 
\label{sec:results}

In Table \ref{tab:results} we provide the fact verification results of the state-of-the-art DA model trained with each of our masking approaches, as well as with the original fully lexicalized input.  
First, we note that indeed the fully lexicalized model, which performs well in-domain, transfers very poorly to a new domain.  
For example, the lexicalized model trained on FEVER,  gave an accuracy of 83.43\% when tested on FEVER, but reduced to 48.86\% when tested on FNC.
This verifies our findings that the signal the model learns from unmasked text does not generalize well.
Additionally, the deletion baseline performs even worse than the fully lexicalized model, which indicates that while the original text can be too domain-specific, its semantic content needs to be maintained on some level.

In terms of this work, we see that all of our masking approaches improve accuracy in the cross-domain setting, by as much as 10.6\% (25.8\% relative), while still maintaining strong in-domain performance (dropping only a few percentage points).
The best cross-domain performance is obtained using our methods which are overlap-aware, suggesting that even when content is abstracted (i.e., masked), the model benefits from explicit awareness of lexical overlap for fact verification. For example, the overlap-aware SS tagging model that was trained on FNC, gave the highest accuracy of 51.77\% when tested out-of-domain, on FEVER.

We note that the model trained on FNC is able to get optimal performance in the FEVER dataset when using the OA-NER + SS tag masking, while the addition of super sense tags did not benefit the model that was trained on FEVER.  
This could be due to the fact that overall, the evidence provided in FNC is much longer than that of FEVER,
and perhaps the model needs more training data to learn stable signal from the SS tags. 
More importantly, this also suggests that while we have demonstrated that semantic masking is clearly beneficial, it is unclear exactly what level of masking granularity should be employed for a given domain -- from coarse-grained NER tags, to the more granular super sense tags, perhaps even to very fine-grained tags such as those proposed by \citet{ling2012fine}.

\section{Conclusion}

We investigated the importance placed by neural network methods for textual entailment on various lexical items. We concluded that attention weights tend to be directed towards words that are more likely to be domain specific, which 
considerably impacts performance in an out-of-domain setting. 
To mitigate this issue, we introduced several strategies for semantically masking word classes such as nouns and verbs, by generalizing them to more abstract concepts, which are more likely to have similar usage across domains. 
We demonstrated the utility of our approach in a cross-domain evaluation of textual entailment for fact verification. Our approach outperforms a model trained on the original, fully-lexicalized texts by over 10\% accuracy when evaluated out of domain. 
Since our approach is implemented as a data pre-processing step, it can potentially help any neural method that learns from text and is likely to be used out of domain.

\section*{Acknowledgments}

This work was supported by the Defense Advanced Research Projects Agency
(DARPA) under the World Modelers program, grant number
W911NF1810014, and by the Bill and Melinda
Gates Foundation HBGDki Initiative.
Mihai Surdeanu declares a financial interest in lum.ai. This interest has been properly disclosed to the University of Arizona Institutional Review Committee and is managed in accordance with its conflict of interest policies.

\bibliography{emnlp-ijcnlp-2019}
\bibliographystyle{acl_natbib}

\end{document}